\documentclass[10pt,twocolumn,letterpaper]{article}

\usepackage{times}
\usepackage{epsfig}
\usepackage{graphicx}
\usepackage{amsmath}
\usepackage{amssymb}
\usepackage{color,xcolor}

\usepackage{comment}

\usepackage{pifont}

\usepackage{microtype}
\usepackage[font=small]{caption}
\usepackage{arydshln}
\usepackage{tabularx}
\usepackage{adjustbox}
\usepackage{array}
\usepackage{booktabs}
\usepackage{colortbl}
\usepackage{float,wrapfig}
\usepackage{hhline}
\usepackage{multirow}
\usepackage{subcaption} %
\usepackage[percent]{overpic}

\usepackage{stmaryrd}

\usepackage{amsmath,amsfonts,amssymb}
\usepackage{bm}
\usepackage{nicefrac}
\usepackage{microtype}
\usepackage{dsfont}
\usepackage{changepage}
\usepackage{extramarks}
\usepackage{fancyhdr}
\usepackage{setspace}
\usepackage{soul}
\usepackage{xspace}

\usepackage{algorithm, algorithmic}
\usepackage{enumitem}
\usepackage[title]{appendix}

\usepackage{fp}
\usepackage{expl3}[2012-07-08]
\ExplSyntaxOn
\ExplSyntaxOff

\usepackage{amsmath,amsfonts,bm}

\def\x{{x}}

\def\xi{{\x_i}}

\newcommand{\reffig}[1]{Figure~\ref{fig:#1}}
\newcommand{\refsec}[1]{Section~\ref{sec:#1}}

\newcommand{\reftbl}[1]{Table~\ref{tbl:#1}}

\newcommand{\lblfig}[1]{\label{fig:#1}}
\newcommand{\lblsec}[1]{\label{sec:#1}}

\newcommand{\lbltbl}[1]{\label{tbl:#1}}

\newcommand{\ignorethis}[1]{}

\newcommand{\myparagraph}[1]{\smallskip \noindent \textbf{#1}}

\def\eqref#1{equation~\ref{#1}}

\def\1{\bm{1}}

\DeclareMathAlphabet{\mathsfit}{\encodingdefault}{\sfdefault}{m}{sl}
\SetMathAlphabet{\mathsfit}{bold}{\encodingdefault}{\sfdefault}{bx}{n}

\newcommand{\ignore}[1]{}

\makeatletter
\DeclareRobustCommand\onedot{\futurelet\@let@token\@onedot}
\def\@onedot{\ifx\@let@token.\else.\null\fi\xspace}

\def\etal{\emph{et al}\onedot}
\makeatother

\definecolor{MyDarkBlue}{rgb}{0,0.08,1}
\definecolor{MyDarkGreen}{rgb}{0.02,0.6,0.02}
\definecolor{MyDarkRed}{rgb}{0.8,0.02,0.02}
\definecolor{MyDarkOrange}{rgb}{0.40,0.2,0.02}
\definecolor{MyPurple}{RGB}{111,0,255}
\definecolor{MyRed}{rgb}{1.0,0.0,0.0}
\definecolor{MyGold}{rgb}{0.75,0.6,0.12}
\definecolor{MyDarkgray}{rgb}{0.66, 0.66, 0.66}
\definecolor{myorange}{RGB}{255,69,0}

\usepackage{cvpr}              %

\usepackage{graphicx}
\usepackage{amsmath}
\usepackage{amssymb}
\usepackage{booktabs}
\usepackage{microtype}

\usepackage[pagebackref,breaklinks,colorlinks]{hyperref}

\usepackage[capitalize]{cleveref}
\crefname{section}{Sec.}{Secs.}
\Crefname{section}{Section}{Sections}
\Crefname{table}{Table}{Tables}
\crefname{table}{Tab.}{Tabs.}

\def\cvprPaperID{4858} %
\def\confName{CVPR}
\def\confYear{2022}

\begin{document}

\title{Depth-supervised NeRF:  Fewer Views and Faster Training for Free}

\author{Kangle Deng\textsuperscript{1} \qquad Andrew Liu\textsuperscript{2}  \qquad Jun-Yan Zhu\textsuperscript{1} \qquad Deva Ramanan\textsuperscript{1,3}
\\
\textsuperscript{1}Carnegie Mellon University \qquad \textsuperscript{2}Google \qquad \textsuperscript{3}Argo AI }

\twocolumn[{%
\maketitle
\centering
\includegraphics[width=0.95\linewidth]{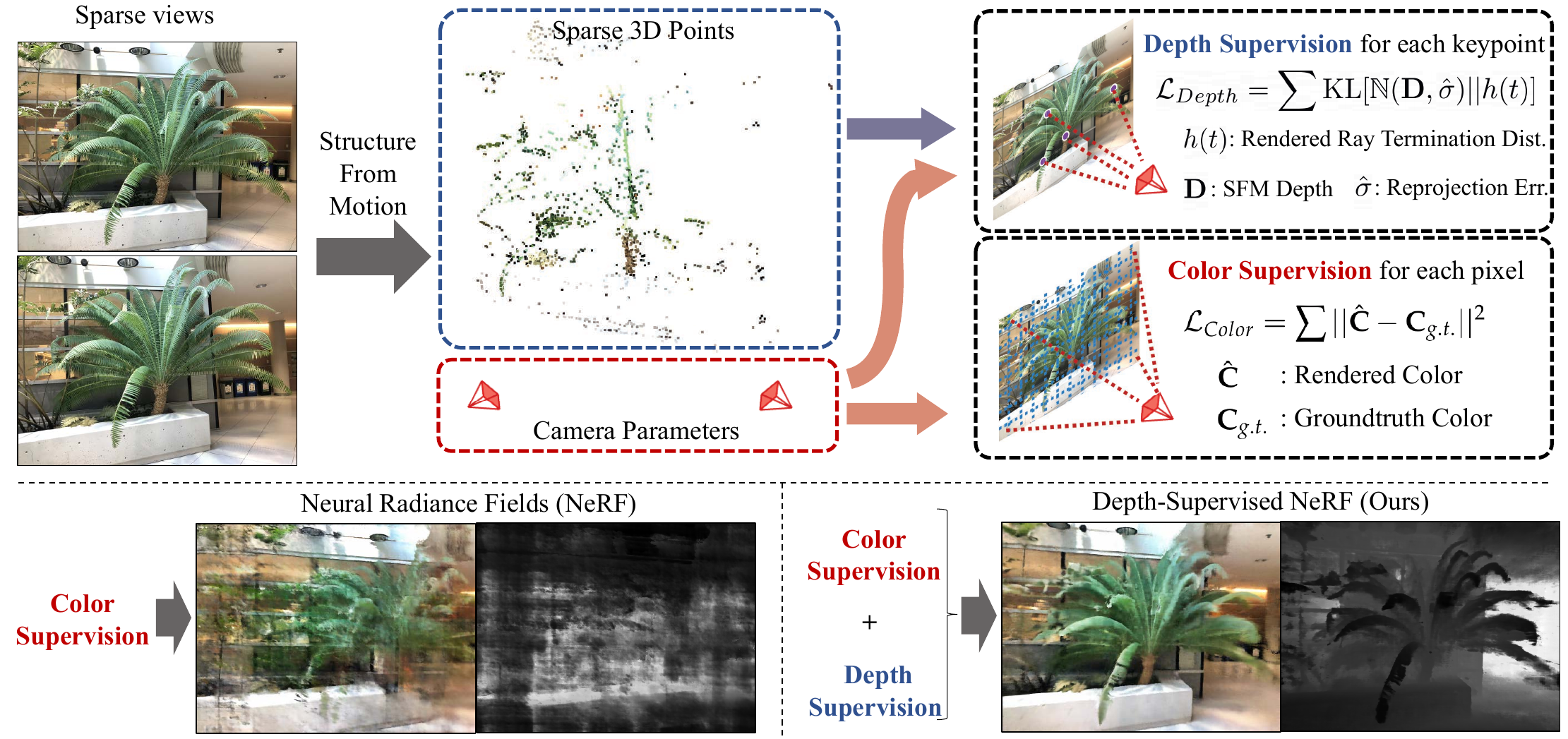}
 \vspace{-0.5em}
\captionof{figure}{
Training NeRFs can be difficult when given insufficient input images. We utilize additional supervision from depth recovered from 3D point clouds estimated from running structure-from-motion and impose a loss to ensure the rendered ray's termination distribution respects the surface priors given by the each keypoint. Because our supervision is complementary to NeRF, it can be combined with any such approach to reduce overfitting and speed up training.
}

 \vspace{1em}
\lblfig{teaser}
}]

\begin{abstract}
A commonly observed failure mode of Neural Radiance Field (NeRF) is fitting incorrect geometries when given an insufficient number of input views. One potential reason is that standard volumetric rendering does not enforce the constraint that most of a scene's geometry consist of empty space and opaque surfaces. We formalize the above assumption through DS-NeRF (Depth-supervised Neural Radiance Fields), a loss for learning radiance fields that takes advantage of readily-available depth supervision. 
We leverage the fact that current NeRF pipelines require images with known camera poses that are typically estimated by running structure-from-motion (SFM). Crucially, SFM also produces sparse 3D points that can be used as ``free" depth supervision during training: we add a loss to encourage the distribution of a ray's terminating depth matches a given 3D keypoint, incorporating depth uncertainty.
DS-NeRF can render better images given fewer training views while training 2-3x faster.
Further, we show that our loss is compatible with other recently proposed NeRF methods, demonstrating that depth is a cheap and easily digestible supervisory signal. 
And finally, we find that DS-NeRF can support other types of depth supervision such as scanned depth sensors and RGB-D reconstruction outputs.

\end{abstract}

\section{Introduction}
\begin{figure*}[h]
    \centering
    \includegraphics[width=\linewidth]{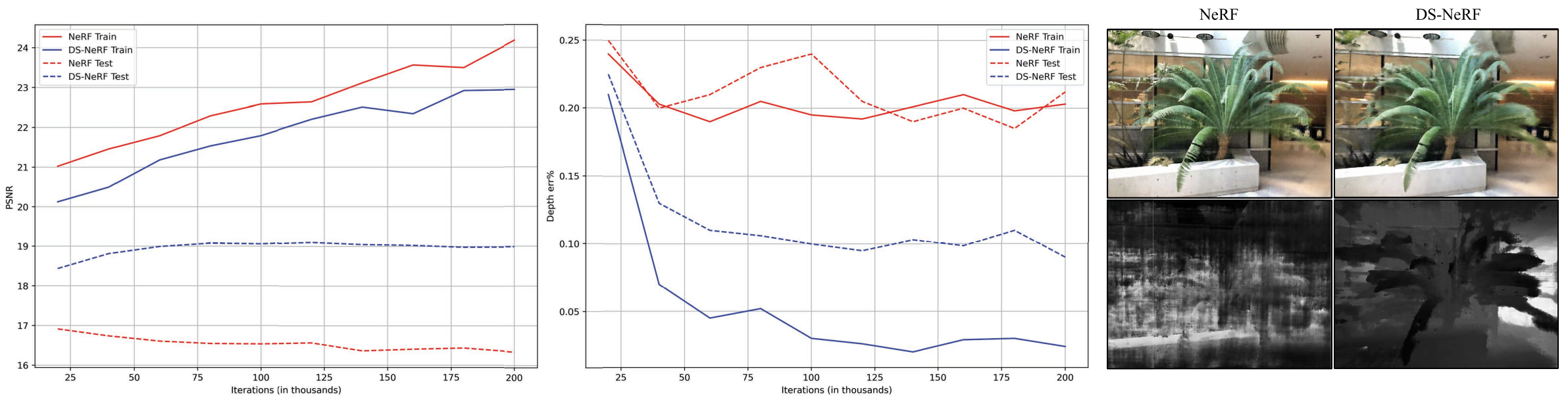}
    \vspace{-1.5em}
    \caption{\textbf{Few view NeRF.} 
NeRF is susceptible to overfitting when given few training views. As seen by the PSNR gap between train and test renders (left), NeRF has overfit and fails at synthesizing novel views. Further, the depth map (right) and depth error (middle) for NeRF suggest that its density function has failed to extract the surface geometry and can only reconstruct the training views' colors. Our depth-supervised NeRF model is able to render plausible geometry with consistently lower depth errors.}
\lblfig{nerf_analysis}
\vspace{-0.5em}
\end{figure*}

Neural rendering with implicit representations has become a widely-used technique for solving many vision and graphics tasks ranging from view synthesis~\cite{sitzmann2019srns,mildenhall2020nerf, SRF}, to re-lighting~\cite{Meshry_2019_CVPR,martinbrualla2020nerfw},  to pose and shape estimation~\cite{Park_2019_CVPR,Saito_2019_ICCV,yen2020inerf}, to 3D-aware image synthesis and editing~\cite{schwarz2020graf,chan2020pi,liu2021editing}, to modeling dynamic scenes~\cite{park2020nerfies,pumarola2020d,li2020neural}. %
The seminal work of Neural Radiance Fields (NeRF)~\cite{mildenhall2020nerf} demonstrated impressive view synthesis results by using implicit functions to encode volumetric density and color observations. %

In spite of this, NeRF has several limitations. Reconstructing both the scene appearance and geometry can be ill-posed given a small number of input views. \reffig{nerf_analysis} shows that NeRF can learn wildly inaccurate scene geometries that still accurately render train-views. %
 However, such models produce poor renderings of novel test-views, essentially overfitting to the train set. %
Furthermore, even given a large number of input views, NeRF can still be time-consuming to train;
it often takes between ten hours to several days to model a single scene at moderate resolutions on a single GPU. The training is slow due to both the expensive ray-casting operations and lengthy optimization process.

In this work, we explore depth as an additional, cheap source of supervision to guide the geometry learned by NeRF. Typical NeRF pipelines require images and camera poses, where the latter are estimated from structure-from-motion (SFM) solvers such as COLMAP~\cite{schoenberger2016sfm}. In addition to returning cameras, COLMAP also outputs sparse 3D point clouds as well as their reprojection errors. %
We impose a loss to encourage the distribution of a ray's termination to match the 3D keypoint, incorporating reprojection error as an uncertainty measure. This is a significantly stronger signal than reconstructing only RGB. Without depth supervision, NeRF is implicitly solving a 3D correspondence problem between multiple views. However, the sparse version of this exact problem has already been solved by SFM, whose solution is given by the sparse 3D keypoints. Therefore depth supervision improves NeRF by (softly) anchoring its search over implicit correspondences with sparse explicit ones.

Our experiments show that this simple idea translates to massive improvements in training NeRFs and its variations, regarding both the training speed and the amount of training data needed. We observe that depth-supervised NeRF can accelerate model training by 2-3x while producing results with the same quality. For sparse view settings, experiments show that our method synthesizes better results compared to the original NeRF and recent sparse-views NeRF models~\cite{yu2020pixelnerf,tancik2020meta} on both NeRF Real~\cite{mildenhall2020nerf} and Redwood-3dscan~\cite{Choi2016redwood}
We also show that our depth supervision loss works well with depth derived from other sources such as a depth camera. 
Our code and more results are available at \url{https://www.cs.cmu.edu/~dsnerf/}. %

\section{Related Work}

\myparagraph{NeRF from few views.}
NeRF~\cite{mildenhall2020nerf} was originally shown to work on a large number of images with the LLFF NeRF Real dataset~\cite{mildenhall2019llff} consisting of nearly 50 images per scene. This is because fitting the NeRF volume often requires a large number of views to avoid arriving at degenerate representations. Recent works have sought to decrease the data-hungriness of NeRF in a variety of different ways. PixelNeRF~\cite{yu2020pixelnerf} and metaNeRF~\cite{tancik2020meta} use data-driven priors recovered from a domain of training scenes to fill in missing information from test scenes. Such an approach works well when given sufficient training scenes and limited gap between the training and test distribution, but such assumptions are not particularly flexible. Another approach is to leverage priors recovered from a different task like semantic consistency~\cite{jain2021putting} or depth prediction~\cite{wei2021nerfingmvs}.

Similar to our insight that the primary difficulty in fitting few view NeRF is correctly modeling 3D geometry, MVSNeRF~\cite{chen2021mvsnerf} combines both 3D knowledge with scene priors by constructing a plane sweep volume before using a pretrained network with generalizable priors to render scenes. One appeal of an approach that utilizes 3D information is the lack of assumption it makes on the problem statement. Unlike the aforementioned approaches which depend on the availability of training data or the applicability of prior assumptions, our approach only requires the existence of 3D keypoints. This gives depth supervision the flexibility to be used not only as a standalone method, but one that can be freely incorporated into existing NeRF methods easily.

\myparagraph{Faster NeRF.}
Another drawback of NeRF is the lengthy optimization time required to fit the volumetric representation. Indeed Mildenhall \etal~\cite{mildenhall2020nerf} trained a single scene's NeRF model for twelve hours of GPU compute. Many works \cite{Reiser2021KiloNeRFSU,yu2021plenoctrees} have found that the limiting factor is not learning the radiance itself, but rather oversampling the empty space during training. Indeed this is a similar intuition to the fact that the majority of the volume is actually empty, but NeRF's initialization is a median uniform density. Our insight is to apply a supervisory signal directly to the NeRF density to increase the convergence of the geometry and to encourage NeRF's density function to mimic the behavior of real world surface geometries. 

\myparagraph{Depth and NeRF.}
Several prior works have explored ways to leverage depth information for view synthesis~\cite{single_view_mpi,Shih3DP20} and NeRF training \cite{liu2020neural,park2020nerfies,li2020neural,neff2021donerf,wei2021nerfingmvs}. For instance, 3D keypoints have been demonstrated to be helpful when extending NeRFs with relaxed assumptions like deformable surfaces~\cite{park2020nerfies} or dynamic scene flows~\cite{li2020neural}. Other works like DONeRF~\cite{neff2021donerf} proposed training a depth oracle to improve rendering speed by directly smartly sampling the surface of a NeRF density function. Similar to DONeRF, NerfingMVS~\cite{wei2021nerfingmvs} shows how a monocular depth network can be used to induce depth priors to do smarter sampling during training and inference.

Our work attempts to improve NeRF-based methods by directly supervising the NeRF density function. As depth becomes a more accessible source of data, being able to apply depth supervision becomes increasingly more powerful. For example, recent works have demonstrated how depth extracted from sensors like time-of-flight cameras~\cite{attal2021torf} or RGB-D Kinect sensor~\cite{azinovic2021neural} can be applied to fit implicit functions. 
Building upon their insights, we provide a probabilistic formulation of the depth supervision, and show this results in meaningful improvements to NeRF and its variants.

\section{Depth-Supervised Ray Termination}
\lblsec{method}

\begin{figure*}[h]
    \centering
    \includegraphics[width=\linewidth]{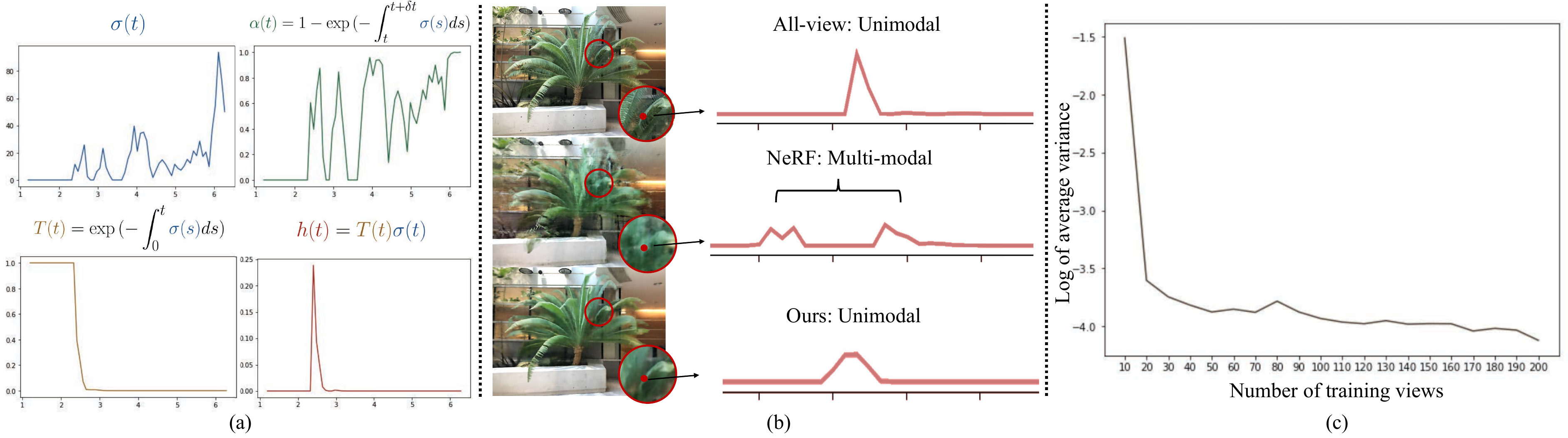}
    \vspace{-1.8em}
    \caption{
\textbf{Ray Termination Distribution.} (a) We plot various NeRF components over the distance traveled by the ray. Even if a ray traverses through multiple objects (as indicated by the multiple peaks of density $\sigma(t)$), we find that the ray termination distribution $h(t)$ is still unimodal. We find that NeRF models trained with sufficient supervision tend to have peaky, unimodal ray termination distributions as seen by the decreasing variance with more views in (c). We posit that the ideal ray termination distribution approaches a $\delta$ impulse function.} 

\lblfig{raydistribution}
\vspace{-0.5em}
\end{figure*}

We now present our proposed depth-supervised loss for training NeRFs. We first revisit volumetric rendering and then analyze the termination distribution for rays. We conclude with our depth-supervised distribution loss.

\subsection{Volumetric rendering revisited}

A Neural Radiance Field takes a set of posed images and encodes a scene as a volume density and emitted radiance. More specifically, for a given 3D point $\textbf{x} \in \mathbb{R}^3$ and a particular viewing direction $\textbf{d} \in \mathbb{R}^3$, NeRF learns an implicit function $f$ that estimates the differential density $\sigma$ and RGB color $\textbf{c}$ like so: $f(\textbf{x}, \textbf{d}) = (\sigma, \textbf{c})$. 

To render a 2D image given a pose $\mathbf{P}$, we cast rays $\mathbf{r}$ originating from the $\mathbf{P}$'s center of projection $\mathbf{o}$ in direction $\mathbf{d}$ derived from its intrinsics. We integrate the implicit radiance field along this ray to compute the incoming radiance from any object that lies along $\mathbf{d}$:
\begin{equation} 
 \hat{\mathbf{C}} = \int_{0}^{\infty} T(t)\sigma(t)\mathbf{c}(t) dt,
\label{eq:color_render}
\end{equation}
where $t$ parameterizes the aforementioned ray as $\mathbf{r}(t) = \mathbf{o} + t\mathbf{d}$ and $T(t) = \exp(-\int_{0}^t \sigma(s) ds)$ checks for occlusions by integrating the differential density between $0$ to $t$. Because the density and radiance are the outputs of neural networks, NeRF methods approximate this integral using a sampling-based Riemann sum instead. The final NeRF rendering loss is given by a reconstruction loss over colors returned from rendering the set of rays $\mathcal{R}(\mathbf{P})$ produced by a particular camera parameter $\mathbf{P}$.

\begin{equation} 
 \mathcal{L}_{\mathrm{Color}} = \mathbb{E}_{\mathbf{r} \in \mathcal{R}(\mathbf{P)}} \left\lVert\hat{\mathbf{C}}(\mathbf{r}) - \mathbf{C}(\mathbf{r})\right\rVert_2^2.
\end{equation}

\myparagraph{Ray distribution.}
Let us write $h(t) = T(t)\sigma(t)$. %
In the appendix, we show that it is a continuous probability distribution over ray distance $t$ that describes the likelihood of a ray terminating at $t$. Due to practical constraints, NeRFs %
assume that the scene lies between a near and far bound ($t_n$, $t_f$). To ensure $h(t)$ sums to one, NeRF implementations often treat $t_f$ as an opaque wall. With this definition, the rendered color can be written as an expectation: %
\begin{equation*}
    \hat{\mathbf{C}} = \int_{0}^{\infty} h(t)\mathbf{c}(t) dt = \mathbb{E}_{h(t)} [\mathbf{c}(t)].
\end{equation*}

\myparagraph{Idealized distribution.} The distribution $h(t)$ describes the weighed contribution of sampled radiances along a ray to the final rendered value. %
Most scenes consist of empty spaces and opaque surfaces that restrict the weighted contribution to stem from the closest surface. This implies that the ideal ray distribution of image point with a closest-surface depth of $\mathbf{D}$ should be $\delta (t - \mathbf{D})$.
\reffig{raydistribution}(c) shows that the empirical variance of NeRF termination distributions decreases with more training views, suggesting that %
high quality NeRFs (trained with many views) tend to have ray distributions that approach the $\delta$-function. %
This insight motivates our depth-supervised ray termination loss.

\subsection{Deriving depth-supervision}

Recall that most NeRF pipelines require images with associated camera matrices $(\mathbf{P}_1, \mathbf{P}_2, \ldots)$, often estimated with SFM packages such as COLMAP~\cite{schoenberger2016sfm}. Importantly, SFM makes use of bundle adjustment, which also returns 3D keypoints $\{\mathbf{X}:\mathbf{x}_1, \mathbf{x}_2, \ldots \in \mathbb{R}^3\}$ and visibility flags for which keypoints are seen from camera $j$:  $\mathbf{X}_j\subset \mathbf{X}$.
Given image $I_j$ and its camera $\mathbf{P}_j$, we estimate the depth of visible keypoints  $\mathbf{x}_i \in \mathbf{X}_j$ by simply projecting $\mathbf{x}_i$ with $\mathbf{P}_j$, taking the re-projected $z$ value as the keypoint's depth $\mathbf{D}_{ij}$.
 
\myparagraph{Depth uncertainty.} Unsurprisingly $\mathbf{D}_{ij}$ are inherently noisy estimates due to spurious correspondences, noisy camera parameters, or poor COLMAP optimization. The reliability of a particular keypoint $\mathbf{x}_i$ can be measured using the average reprojection error $\hat\sigma_i$ across views over which the keypoint was detected. %
Specifically, we model the location of the first surface encountered by a ray as a random variable $\mathbb{D}_{ij}$ that is normally distributed around the COLMAP-estimated depth $\mathbf{D}_{ij}$ with variance $\hat\sigma_i$:
 $  \mathbb{D}_{ij} \sim \mathbb{N}(\mathbf{D}_{ij}, \hat\sigma_i )$.
Combining the intuition regarding behavior of ideal termination distribution, our objective is to minimize the KL divergence between the rendered ray distribution $h_{ij}(t)$ of $\mathbf{x}_i$'s image coordinates and the noisy depth distribution:
\begin{small}
\begin{align}
      \mathbb{E}_{\mathbb{D}_{ij}}   \text{KL}[\delta (t - \mathbb{D}_{ij}) |\vert h_{ij}(t)] = \text{KL}[\mathbb{N}(\mathbf{D}_{ij}, \hat\sigma_i ) |\vert h_{ij}(t)] + const. \nonumber
\end{align}
\end{small}

\noindent {\bf Ray distribution loss.} The above equivalence (see our appendix for proof) %
allows the termination distribution $h(t)$ to be trained with probabilisitic COLMAP depth supervision:

\begin{small}
\begin{equation*} 
\begin{aligned}
\mathcal{L}_{Depth}  %
                    &=  \mathbb{E}_{x_i \in X_j} \left[ - \int \log h(t) \exp{(-\frac{(t-\mathbf{D}_{ij})^2}{2\hat\sigma_i^2})} dt \right] \\
                    &\approx \mathbb{E}_{x_i \in X_j} \left[ - \sum_{k} \log h_k\exp{(-\frac{(t_k-\mathbf{D}_{ij})^2}{2\hat\sigma_i^2})}\Delta t_k \right].
\end{aligned}
\end{equation*}
\end{small}

\noindent Our overall training loss for NeRF is
$    \mathcal{L} = \mathcal{L}_{\mathrm{Color}} + \lambda_{D} \mathcal{L}_{\mathrm{Depth}} $
where $\lambda_{D}$ is a hyper-parameter balancing color and depth supervision.

\section{Experiments}
\lblsec{expr}

\begin{figure*}[t]
    \centering
    \includegraphics[width=\textwidth]{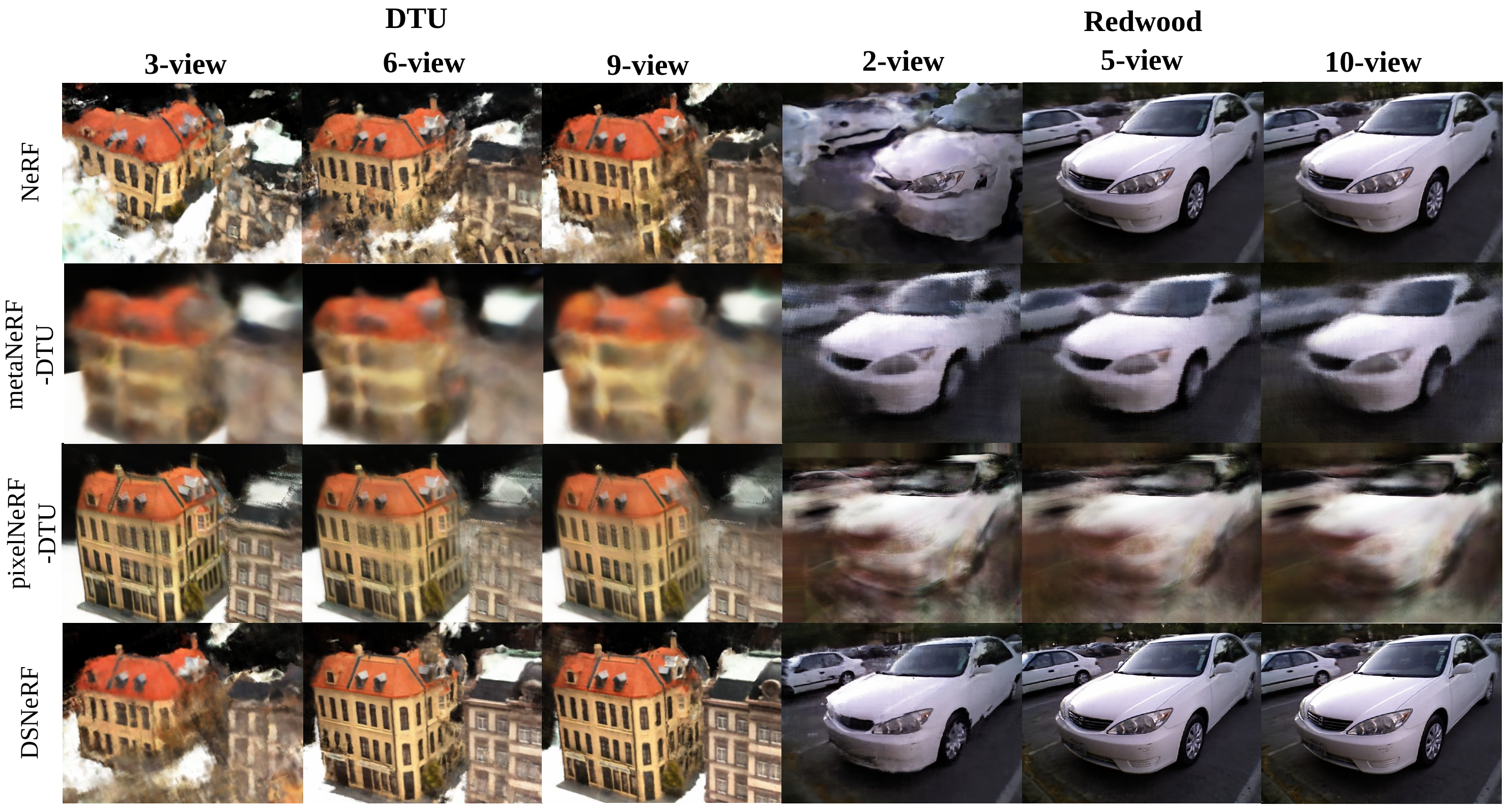}
    \vspace{-10pt}
    \caption{\textbf{View Synthesis on DTU and Redwood.} PixelNeRF, which is pre-trained on DTU, performs the best when given 3-views, although we find DS-NeRF to be visually competitive when more views are available. On Redwood, DS-NeRF is the only baseline to perform well on the 2-views setting.}
    \label{fig:expB}
    \vspace{-0.5em}
\end{figure*}
\begin{figure*}[t]
    \centering
    \includegraphics[width=\textwidth]{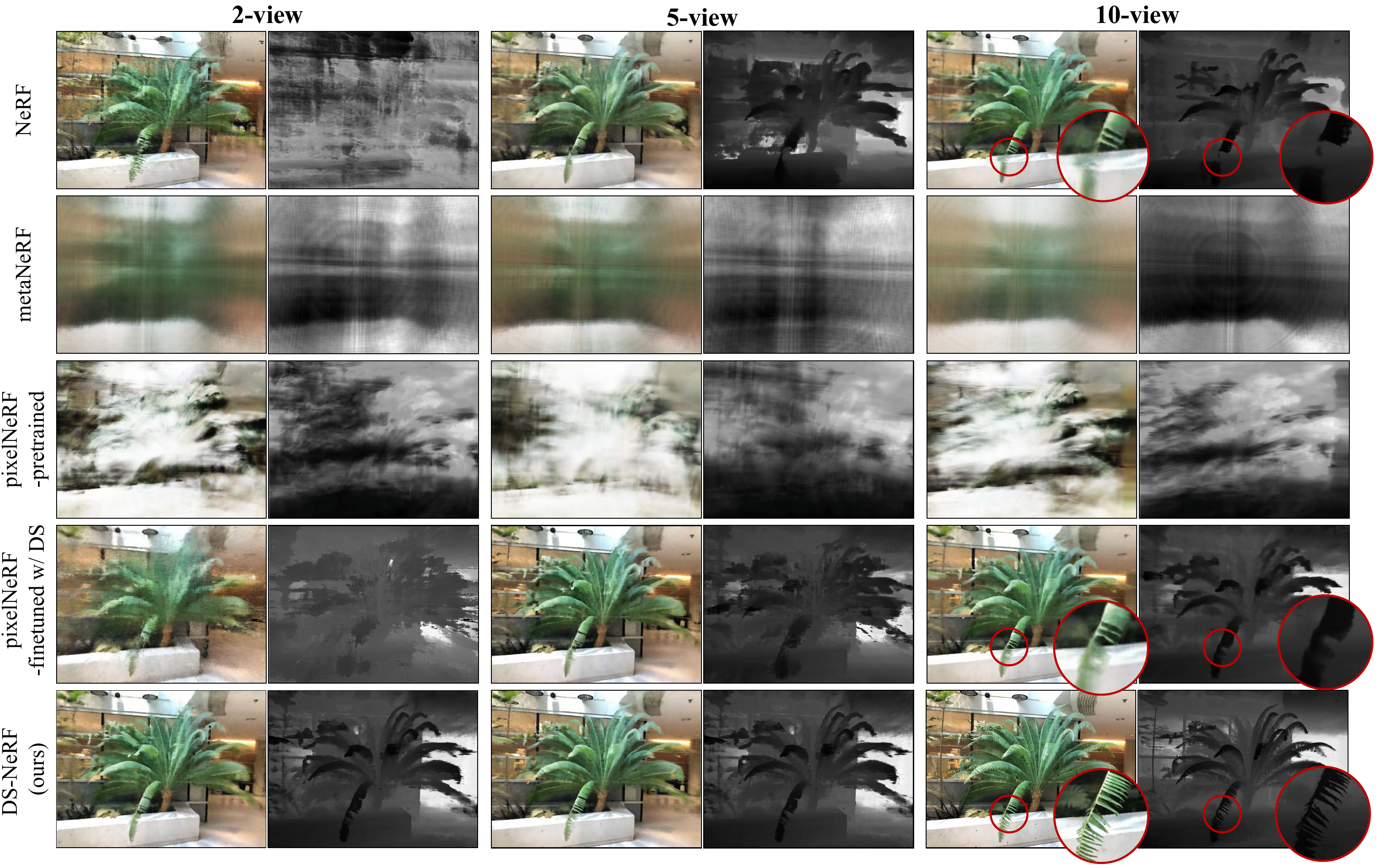}
    \vspace{-1.5em}
    \caption{\textbf{Qualitative Comparison on NeRF Real.} We render novel views and depth for various NeRF models trained on 2, 5, and 10 views. We find that methods trained with DTU struggle on NeRF Real while methods that use depth-supervision are able to render test views with realistic depth maps, even when only 2 views are provided. Please refer to \reftbl{nerf_real} for quantitative comparisons.}
    \lblfig{comp_nerf}
    \vspace{-0.5em}
\end{figure*}

\begin{figure*}[t]
    \centering
    \includegraphics[width=\textwidth]{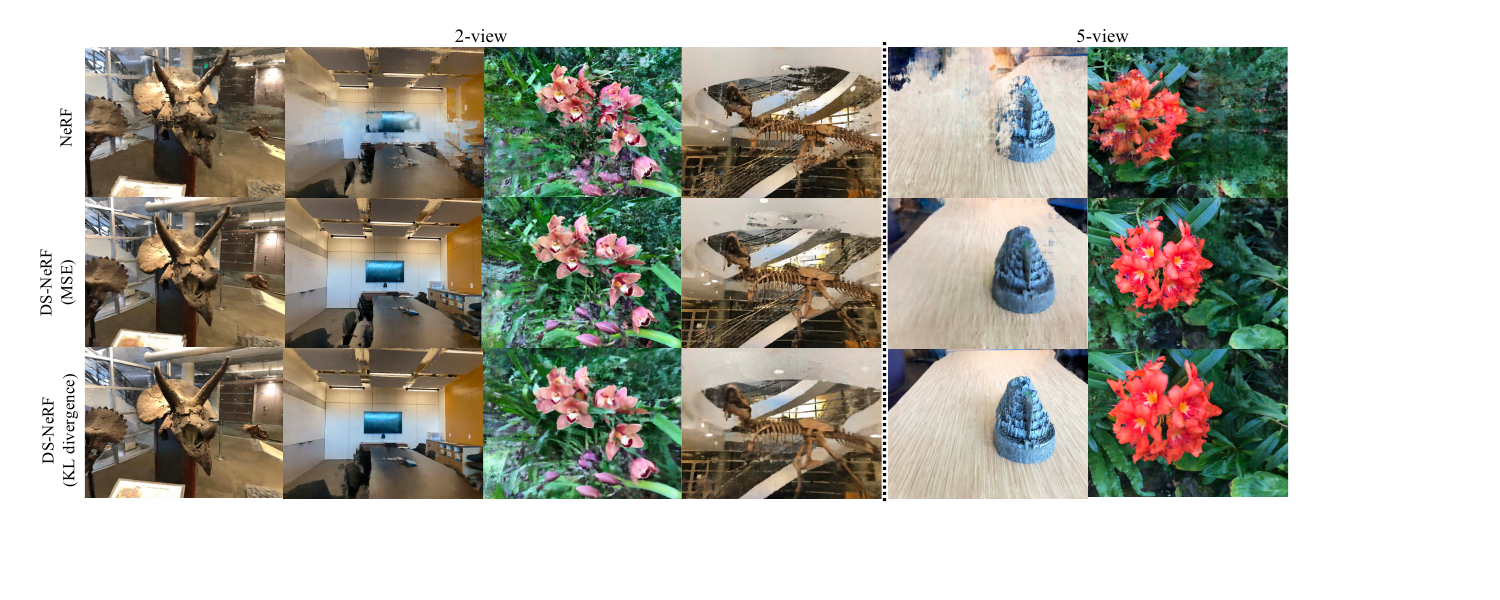}
    \vspace{-1em}
    \caption{\textbf{Depth Supervision Ablations.} We render novel views for NeRF and DS-NeRF trained on 2 views and 5 views. NeRF fails to render novel views as evident by the many artifacts. Using MSE between rendered and sparse depth improves results slightly, but with KL Divergence, DS-NeRF is able to render images with the fewest artifacts.}
    \lblfig{abl}
    \vspace{-1em}
\end{figure*}

\begin{table*}
\centering
\resizebox{\linewidth}{!}{
\setlength{\tabcolsep}{8.5pt}
\begin{tabular}{@{}l |  c c c | c c c | c c c }
\toprule
 & \multicolumn{3}{c}{PSNR$\uparrow$} & \multicolumn{3}{c}{SSIM$\uparrow$} & \multicolumn{3}{c}{LPIPS$\downarrow$}  \\
 \textbf{NeRF Real}~\cite{mildenhall2019llff} & 2-view & 5-view & 10-view & 2-view & 5-view & 10-view & 2-view & 5-view & 10-view\\
\midrule
LLFF                     & 14.3 & 17.6 & 22.3 & 0.48 & 0.49 & 0.53 & 0.55 & 0.51 & 0.53 \\
NeRF                     & 13.5 & 18.2 & 22.5 & 0.39 & 0.57 & 0.67 & 0.56 & 0.50 & 0.52\\
metaNeRF-DTU             & 13.1 & 13.8 & 14.3 & 0.43 & 0.45 & 0.46 & 0.89 & 0.88 & 0.87 \\
pixelNeRF-DTU            & 9.6 & 9.5 & 9.7 & 0.39 & 0.40 & 0.40 & 0.82 & 0.87 & 0.81 \\
\quad finetuned          & 18.2 & 22.0 & 24.1 & 0.56 & 0.59 & 0.63 & 0.53 & 0.53 & 0.41  \\
\quad finetuned w/ DS  & 18.9 & 22.1 & 24.4 & 0.54 & 0.61 & 0.66 & 0.55 & 0.47 & 0.42 \\
IBRNet                   & 14.4 & 21.8 & 24.3 & 0.50 & 0.51 & 0.54 & 0.53 & 0.54 & 0.51 \\
\quad finetuned w/ DS  & 19.3 & 22.3 & 24.5 & 0.63 & 0.66 & 0.68 & \textbf{0.39} & 0.36 & 0.38 \\ 
MVSNeRF                  &  -   & 17.2 & 17.2 & -    & 0.61 & 0.60 &    - & 0.37 & 0.36 \\
\quad fintuned           &  -   & 21.8 & 22.9 & -    & \textbf{0.70} & 0.74 & - & \textbf{0.27} & \textbf{0.23} \\
\quad fintuned w/ DS           &  -   & 22.0 & 22.9 & -    & \textbf{0.70} & \textbf{0.75} & - & 0.27 & 0.25 \\
\midrule
DS-NeRF  \\ %
\quad MSE & 19.5 & 22.2 & 24.7  & 0.65 & 0.69 & 0.71 & 0.43 & 0.40 & 0.37 \\
\quad KL divergence & \textbf{20.2} & \textbf{22.6} & \textbf{24.9} & \textbf{0.67} & 0.69  & 0.72 & \textbf{0.39} & 0.35 & 0.34 \\
\bottomrule
\end{tabular}
}
\vspace{-0.7em}
\caption{\textbf{View Synthesis on NeRF Real.} We evaluate view synthesis quality for various methods when given 2, 5, 10 views from NeRF Real. We find that metaNeRF-DTU and pixelNeRF-DTU struggle to learn on NeRF Real due to its domain gap to DTU. PixelNeRF, IBRNet and MVSNeRF can benefit from incorporating the depth supervision loss to achieve their best performance.  We find that our DS-NeRF outperforms these methods on a variety of metrics, but especially for the few view settings like 2 and 5 views.
}
\lbltbl{nerf_real}
\end{table*}

\begin{figure*}[t]
    \centering
    \includegraphics[width=.9\textwidth]{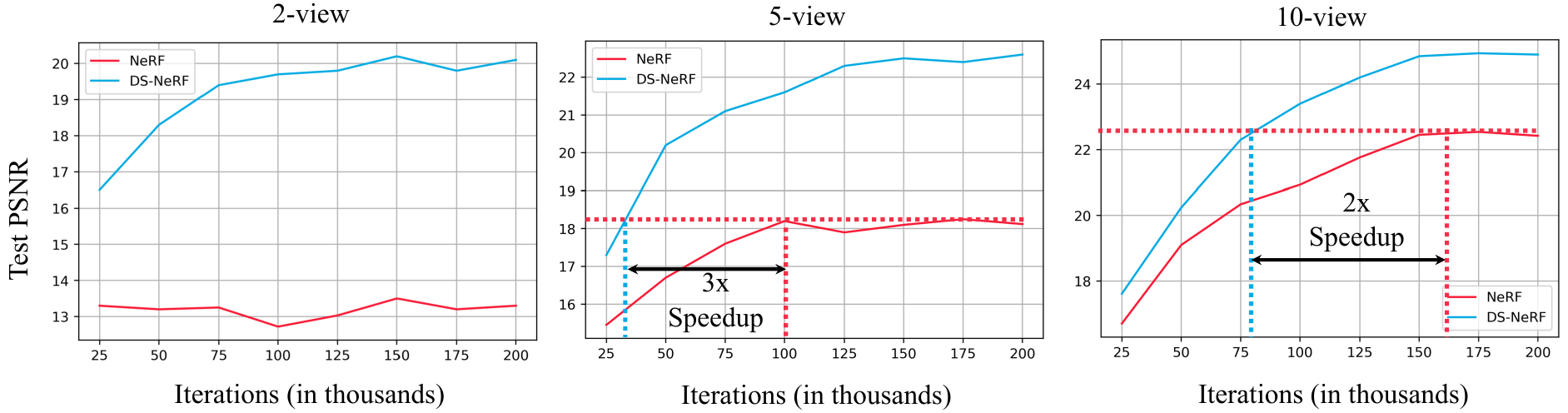}
    \vspace{-0.5em}
    \caption{\textbf{Faster Training.} We train DS-NeRF and NeRF under identical conditions and observe that DS-NeRF is able to reach NeRF's peak PSNR quality in a fraction of the number of iterations across. For 2 views, we find that NeRF is unable to match DS-NeRF's performance.} %
    \lblfig{efficiency}
    \vspace{-1em}
\end{figure*}

\begin{table*}
\setlength{\tabcolsep}{8.5pt}
\center
\begin{tabular}{@{}l |  c c c | c c c | c c c}
\toprule
 & \multicolumn{3}{c}{PSNR$\uparrow$} & \multicolumn{3}{c}{SSIM$\uparrow$} & \multicolumn{3}{c}{LPIPS$\downarrow$}  \\
 \textbf{DTU}~\cite{jensen2014dtu} & 3-view & 6-view & 9-view & 3-view & 6-view & 9-view & 3-view & 6-view & 9-view\\
\midrule
NeRF & 9.9 & 18.6 & 22.1 & 0.37 & 0.72 & \textbf{0.82} & 0.62 & 0.35 & 0.26  \\ 
metaNeRF-DTU & 18.2 & 18.8 & 20.2 & 0.60 & 0.61 & 0.67 & 0.40 & 0.41 & 0.35 \\
pixelNeRF-DTU  & \textbf{19.3} & 20.4 & 21.1 & \textbf{0.70} & 0.73 & 0.76 & \textbf{0.39} & 0.36 & 0.34 \\

\midrule
DS-NeRF \\
\quad MSE & 16.5 & 20.5 & 22.2 & 0.54 & 0.73 & 0.77 & 0.48 & 0.31 & 0.26 \\
\quad KL divergence & 16.9 & \textbf{20.6} & \textbf{22.3}  & 0.57 & \textbf{0.75} & 0.81 & 0.45 & \textbf{0.29} & \textbf{0.24} \\

\bottomrule
\end{tabular}
\caption{\textbf{View Synthesis on DTU.} %
We evaluate on 3, 6, and 9 views respectively for 15 test scenes from the DTU dataset. pixelNeRF-DTU and metaNeRF-DTU perform well given that the domain overlap between training and testing. This is especially true for the few view setting as the lack of information is supplemented by exploiting dataset priors. In spite of this, DS-NeRF is still competitive on view synthesis for 6 and 9 views. 
}
\lbltbl{dtu}
\vspace{-1em}
\end{table*}

\begin{table*}
\setlength{\tabcolsep}{8.2pt}
\center
\begin{tabular}{@{}l |  c c c | c c c | c c c}
\toprule
 & \multicolumn{3}{c}{PSNR$\uparrow$} & \multicolumn{3}{c}{SSIM$\uparrow$} & \multicolumn{3}{c}{LPIPS$\downarrow$}  \\
 \textbf{Redwood-3dscan}~\cite{Choi2016redwood} & 2-view & 5-view & 10-view & 2-view & 5-view & 10-view & 2-view & 5-view & 10-view\\
\midrule
NeRF & 10.5 & 22.4 & 23.4 & 0.38 & 0.75 & 0.82 & 0.51 & 0.45 & 0.45 \\ 
metaNeRF-DTU & 14.3 & 14.6 & 15.1 & 0.37 & 0.39 & 0.40 & 0.76 & 0.76 & 0.75 \\ 
pixelNeRF-DTU & 12.7 & 12.9 & 12.8 & 0.43 & 0.47 & 0.50 & 0.76 & 0.75 & 0.70 \\
MVSNeRF-DTU &  - & 17.1 & 17.1 & - & 0.54 & 0.53 & - & 0.63 & 0.63 \\
\quad finetuned & - & 22.7 & 23.1 & - & 0.78 & 0.78 & - & 0.36 & 0.34  \\
\midrule
DS-NeRF  & 18.1  & 22.9  & 23.8  &  0.62 & \textbf{0.78} & 0.81 & 0.40 & \textbf{0.34} & 0.42\\
\midrule
DS-NeRF w/ RGB-D  & \textbf{20.3} & \textbf{23.4} & \textbf{23.9} &\textbf{ 0.73} & 0.77 & \textbf{0.84} & \textbf{0.36} & 0.35 & \textbf{0.28}\\

\bottomrule
\end{tabular}
\caption{\textbf{View Synthesis on Redwood.} We evaluate view synthesis on 2, 5, and 10 input views on the Redwood dataset. DS-NeRF (with COLMAP~\cite{schoenberger2016sfm} inputs) outperforms baselines on various metrics across varying numbers of views. Learning DS-NeRF with the RGB-D reconstruction output~\cite{zeng20173dmatch} further improves performance, highlighting the potential of applying our method alongside other sources of depth. 
 }
\lbltbl{redwood}
    \vspace{-0.2em}
\end{table*}

We first evaluate the input data efficiency on view synthesis over several datasets in \refsec{exp_svs}. For relevant NeRF-related methods, we also evaluate the error of rendered depth maps in \refsec{exp_rdm}. Finally, we analyze training speed improvements in \refsec{exp_te}.

\subsection{Datasets} \label{sec:data}

\myparagraph{DTU MVS Dataset (DTU)}~\cite{jensen2014dtu} captures various objects from multiple viewpoints. Following Yu \etal's setup in PixelNeRF~\cite{yu2020pixelnerf}, we evaluated on the same test scenes and views. For each scene, we used their subsets of size 3, 6, 9 training views. We run COLMAP with the ground truth calibrated camera poses to get keypoints. Images are down-sampled to a resolution of $400 \times 300$ for training and evaluation. 

\myparagraph{NeRF Real-world Data (NeRF Real)}~\cite{mildenhall2019llff,mildenhall2020nerf} contains 8 real world scenes captured from many forward-facing views. We create subsets of training images for each scene of sizes 2, 5, and 10 views. For every subset, we run COLMAP~\cite{schoenberger2016sfm} over its training images to estimate cameras and collect sparse keypoints for depth supervision.

\myparagraph{Redwood-3dscan (Redwood)}~\cite{Choi2016redwood} contains RGB-D videos of various objects. We select 5 RGB-D sequences and create subsets of 2, 5, and 10 training frames for each object. We run COLMAP to get their camera poses and sparse point clouds.
To connect the scale of COLMAP's pose with the scanned depth, we solve a least-squares that best fits detected keypoints to the scanned depth value. Please refer to our appendix for full details.

\subsection{Comparisons}
First we consider Local Lightfield Fusion (\textit{LLFF})~\cite{mildenhall2019llff}, an MPI-based representation that learns from multiple view points. Next we consider a set of NeRF baselines.

\myparagraph{PixelNeRF}~\cite{yu2020pixelnerf} expands upon NeRF by using an encoder to train a general model across multiple scenes. \textit{pixelNeRF-DTU} is evaluated using the released DTU checkpoint. For cases where the train and test domain are different, we finetune using RGB supervision for additional iterations on each test scene to get \textit{pixelNeRF finetuned}.

\myparagraph{MetaNeRF}~\cite{tancik2020meta} finds a better NeRF initialization over a domain of training scenes before running test-time optimization on new scenes. Because DTU is the only dataset large enough for meta-learning, we only consider the \textit{metaNeRF-DTU} baseline which learns an initialization over DTU for $40\mathrm{K}$ meta-iterations and then finetunes for $1000$ steps on new scenes. We follow metaNeRF's ShapeNet experiments to demonstrate its susceptibility to differences between training and testing domains. 

\myparagraph{IBRNet}~\cite{wang2021ibrnet} extends NeRF by using a MLP and ray transformer to estimate radiance and volume density.

\myparagraph{MVSNeRF}~\cite{chen2021mvsnerf} initializes a plane sweep volume from 3 views before converting it to a NeRF by a pretrained network. MVSNeRF can be further optimized using RGB supervision. 

\myparagraph{DS-NeRF (Ours).} To illustrate the effectiveness of KL divergence, we include a variant of DS-NeRF with an MSE loss between the SFM-estimated and the rendered depth. \reffig{abl} qualitatively shows that KL divergence penalty produces views with less artifacts on NeRF Real sequences.

\myparagraph{DS with existing methods.} As our DS loss does not require additional annotation or assumptions, our loss can be inserted into many NeRF-based methods. Here, we also incorporate our loss when finetuning pixelNeRF and IBRNet. %

\subsection{Few-input view synthesis} \label{sec:exp_svs}
We start by comparing each method on rendering test views from few inputs. For view synthesis, we report three metrics (PSNR, SSIM~\cite{wang2004image}, and LPIPS~\cite{zhang2018unreasonableLpips}) that evaluate the quality of rendered views against a ground truth.

\myparagraph{DTU.} We show evaluations on DTU in \reftbl{dtu} and qualitative results in \reffig{expB}. We find that DS-NeRF renders images from 6 and 9 input views that are competitive with pixelNeRF-DTU, however metaNeRF-DTU and pixelNeRF-DTU are able to outperform DS-NeRF on 3-views. This is not  particularly surprising as both methods are trained on DTU scenes and therefore can fully leverage dataset priors. 

\myparagraph{NeRF Real.} As seen in \reftbl{nerf_real}, our approach renders images with better scores than than NeRF and LLFF, especially when only two and fives input views are available. We also find that metaNeRF-DTU and pixelNeRF struggle which highlights their apparent weakness. These DTU-pretrained models struggle to perform well outside of DTU. Our full approach is capable of achieving good rendering results because we do not utilize assumptions on the test scene's structure. We also add our depth supervision loss to other methods like pixelNeRF and IBRNet and find their performances improve, showing that many methods can benefit from adding depth supervision. MVSNeRF has an existing geometry prior handled by its PSV-initialization, thus we did not see an improvement from adding depth supervision.

\myparagraph{Redwood.} Like NeRF Real, we find similar improvements in performance across the Redwood dataset in \reftbl{redwood}. Because Redwood includes depth measurements collected with a sensor, we also consider how alternative sources of depth supervision can improve results. We train DS-NeRF, replacing COLMAP supervision with the scaled Redwood depth measurements and find that the denser depth helps even more, achieving a PSNR of 20.3 on 2-views.

\subsection{Depth error}
\lblsec{exp_rdm}
\begin{table}
\centering
\resizebox{\columnwidth}{!}{
\setlength{\tabcolsep}{8.5pt}
\begin{tabular}{@{}l |  c c c | c c c}
\toprule
Depth err\%$\downarrow$ & \multicolumn{3}{c}{\textbf{NeRF real-world}} & \multicolumn{3}{c}{\textbf{Redwood-3dscan}}  \\
  & 2-view & 5-view & 10-view & 2-view & 5-view & 10-view\\
\midrule
NeRF & 20.32 & 15.00 & 12.41 & 25.32 & 24.34 & 21.34  \\ 
metaNeRF-DTU & 22.23 & 22.07 & 22.30 & 20.84 & 21.12 & 20.96\\
pixelNeRF-DTU & 22.12 & 22.09 & 22.06 & 19.46 & 19.87 & 19.54 \\
\midrule
DS-NeRF
 & \textbf{10.41}  & \textbf{8.61} & \textbf{8.15} & 11.42 & 10.43 & 9.43 \\
 DS-NeRF w/ RGBD & - & - & - & \textbf{5.81} & \textbf{5.31} & \textbf{4.22} \\
\bottomrule
\end{tabular}
}
\vspace{-4pt}
\caption{\textbf{Depth Error.}  We compare rendered depth to reference ``ground-truth" depth obtained from NeRF Real and Redwood RGB-D. %
DS-NeRF is able to extract better geometry as indicated by the lower depth errors from test views. We also show DS-NeRF trained with Redwood's dense supervision can significantly improve NeRF's ability to model the underlying geometry.}
\lbltbl{depth}
 \vspace{-1.5em}
\end{table}

We evaluate NeRF's rendered depth by comparing them to reference ``ground truth" depth measurements. For NeRF Real, we use reference depth of test keypoints recovered from running an all-view dense stereo reconstruction. For Redwood ~\cite{Choi2016redwood}, we align their released 3D models with our cameras by running 3dMatch~\cite{zeng20173dmatch} and generate reference depths for each test view. %
Please refer to our arXiv version for more details regarding depth error evaluation. As shown in \reftbl{depth}, DS-NeRF, trained with supervision obtained only from depth in training views, is able to estimate depth more accurately than all the other NeRF models. While this is not particularly surprising, it does highlight the weakness of training NeRFs only using RGB supervision. For example, in \reffig{comp_nerf}, NeRF tends to ignore geometry and fails to produce any coherent depth map. %

\myparagraph{RBG-D inputs.} We consider a variant of depth supervision using RGB-D input from Redwood. 
We derive dense depth map for each training view using 3DMatch~\cite{zeng20173dmatch} with RGB-D input. With dense depth supervision, we can render rays for any pixel in the valid region, and apply our KL depth-supervision loss. As shown in \reftbl{redwood} and \reftbl{depth}, dense depth supervision produces even better-quality images and significantly lower depth errors.

\subsection{Analysis}
\lblsec{exp_te}

\myparagraph{Overfitting.} \reffig{nerf_analysis} shows that NeRF can overfit to a small number of input views by learning degenerate 3D geometries. Adding depth supervision can assist NeRF to disambiguate geometry and render better novel views. %

\myparagraph{Faster Training.} To quantify speed improvements in NeRF training, we compare training DS-NeRF and NeRF under identical settings. Like in \refsec{exp_svs}, we evaluate view synthesis quality on test views under various number of input views from NeRF Real using PSNR. We can compare training speed performance by plotting PSNR on test views versus training iterations in \reffig{efficiency}.

DS-NeRF achieves a particular test PSNR threshold using 2-3x less training iterations than NeRF. These benefits are significantly magnified when given fewer views. In the extreme case of only 2-views, NeRF is completely unable to match DS-NeRF's performance. While these results are given in terms of training iteration, we can translate them into wall time improvements. 
On a single RTX A5000, a training loop of DS-NeRF takes $\sim$ 362.4 ms/iter while NeRF needs $\sim$ 359.8 ms/iter.
Thus in the 5-view case, DS-NeRF achieves NeRF's peak test PSNR around 13 hours faster, a massive improvement considering the negligible cost.

\myparagraph{Discussion.}
\lblsec{discussion}
We introduce Depth-supervised NeRF, a model for learning neural radiance fields that takes advantage of depth supervision. Our model uses ``free" supervision provided by sparse 3D point clouds computed during standard SFM pre-processing steps. This additional supervision has a significant impact; DS-NeRF trains 2-3x faster and produces better results from fewer training views (improving PSNR from 13.5 to 20.2). While recent research has sought to improve NeRF by exploiting priors learned from category-specific training data, our approach requires no training and thus generalizes (in principle) to any scenes on which SFM succeeds. This allows us to integrate depth supervision to many NeRF-based methods and observe significant benefits. Finally, we provide cursory experiments that explore alternate forms of depth supervision such as active depth sensors.
Please see our arXiv version for a discussion on limitations and societal impact of our paper.

\myparagraph{Acknowledgments. } We thank Takuya Narihira, Akio Hayakawa, Sheng-Yu Wang, Richard Tucker, Konstantinos Rematas, and Michaelu Zollhöfer for helpful discussion. We are grateful for the support from Sony Corporation, Singapore DSTA, and the CMU Argo AI Center for Autonomous Vehicle Research.

\clearpage

\newpage
{\small
\bibliographystyle{ieee_fullname}
\bibliography{main}
}

\clearpage
\newpage

\appendix
\section*{Appendix}
We provide additional implementation details, discussions, and experimental results.

\section{Discussion}

{\bf Limitations.} Depth supervision is only as good as the estimates of depth, as such poor SfM or bad depth measurements can result in failure of the optimization process. Next we assume a Gaussian distribution models the uncertainty of the keypoint's location, but such a simplifying assumption is not necessarily true especially for depth derived from other sources.

{\bf Societal Impact.} Depth supervision is a technique which empowers NeRF to operate on a wider range of experimental setups. While novel view synthesis is not synthetic media, it can open the door to abuse when generating trajectories through a scene. There may also be privacy concerns as using ubiquitous sensor technology to better render sharper details of a scene could capture personally identifiable information.

\section{Derivation Details}

\subsection{Derivation of $h(t)$ as a probability distribution}
In Section 3.1, we claim that $h(t)$, which is a function that describes a contribution weight from a particular distance $t$, is a continuous probability distribution over ray termination. We can verify this by proving that $h(t)$ is non-negative and the integral of $h(t)$ over $t$'s domain is equal to 1.

\begin{equation*}
    h(t) = T(t)\sigma(t) = \exp[-\int_{0}^t \sigma(s) ds]\sigma(t)
\end{equation*}

We start by assuming that $\sigma(s)$ is a real-valued, non-negative (typically a ReLU or softplus activation) function describing the differential density $s$ units away from the camera origin. Because $\sigma(\cdot)$ is real-valued, the inner-value integral over $\sigma(s)$ must also be real, therefore $T(t)$ must be non-negative. As a result, $h(t)$ is the product of two non-negative functions and is therefore non-negative for all values of $t$, satisfying the first property of a probability distribution. The next step is to show that the integral of $h(t)$ over the domain of $t$ is 1.

To do this we need to make an additional assumption that for every ray cast in a scene will eventually intersect an opaque object: $\forall a\ge 0 \, \int_a^\infty \sigma(s) ds = \infty$. This is true for most scenes we care about modeling as radiance is emitted by surfaces. 

\begin{equation*}
\begin{aligned}
    \int_{0}^{\infty} h(t) dt & = \int_{0}^{\infty} \exp[-\int_{0}^t \sigma(s) ds]\sigma(t) dt
\end{aligned}
\end{equation*}
Let $u(t) = -\int_{0}^t \sigma(s) ds$. We can compute the derivative $\frac{d u(t)}{dt} = -\sigma(t)$ and rewrite the above equation in terms of $u(t)$ and $\frac{du(t)}{dt}$.

\begin{equation*}
\begin{aligned}
    \int_{0}^{\infty} h(t) dt & = -\int_{0}^{\infty} \exp[u(t)] \frac{du(t)}{dt}dt = \exp[u(t)] \Big|_\infty^0 \\
     & = \exp[u(0)] - \exp[u(\infty)] \\
     &= \exp[0] - \exp[-\infty] = 1
\end{aligned}
\end{equation*}

This shows that under the above assumptions, $h(t)$ is guaranteed to be a probability distribution. Due to practical constraints, NeRFs are unable to sample the volume to infinity and instead assume that the scene lies between a near and far bound. To ensure the above assumption still holds true, NeRF implementations will often treat the furthest radiance as an opaque wall.

\subsection{Depth-supervision implementation}
Depth supervision is implemented by projecting a ray with direction (in local camera coordinates) given by the image coordinates of a detected keypoint and $-1$ in the camera axis. We shoot this ray into a scene and render its depth using the same sampling procedure described in NeRF.

For setups where the training data can fit into GPU memory, the most time-consuming part during training comes from the many forward passes required for a single ray marching rendering step. To gain the benefits of faster training, we must simultaneously train with color supervision and depth supervision with a single ray marching procedure.

To do this, we exploit the fact that image coordinates are actually continuous and the pixels are only samples of the color function at discrete intervals. Therefore we can interpolate RGB supervision for rays corresponding to detected keypoints, allowing us to supervise RGB and depth at the same time. We allocate a portion of the training rays to this.

\subsection{COLMAP details}
We run the COLMAP with the default configuration on the limited views (the same as NeRF training inputs) (e.g., 2 views). %
The SfM output is used only during training and is not required for synthesizing novel view. 

\subsection{metaNeRF and pixelNeRF baselines}
\label{sec:baseline}
\myparagraph{metaNeRF.} We use the metaNeRF implementation released by Tancik \etal~\cite{tancik2020meta} which uses Jax. For meta-initialization comparison, we adopt their ShapeNet experimental setup of training a category-specific initialization and further fine-tuning for a fixed number of iterations. In our case, we treat the entire DTU dataset as a single category. %
We used 64 inner loop optimizations and trained for $40\mathrm{K}$ outer loop steps. We subsequently fine-tune for $1\mathrm{K}$ steps.

For adapting metaNeRF across different domains, we have to deal with different ray bounds and coordinate scales. This is challenging, and we do not directly address this issue as devising approaches to transferring NeRF meta-initialization across different datasets is beyond the scope of this paper. Instead, we used the default scaling and ray bounds provided by previous NeRF works for DTU and NeRF Real~\cite{yu2020pixelnerf,mildenhall2020nerf}.

\myparagraph{pixelNeRF-DTU with finetuning.} We start with Yu \etal~\cite{yu2020pixelnerf}'s pre-trained models on DTU dataset. For a new input scene, we finetune the weights of pixelNeRF on training views for $20\mathrm{K}$ iterations before subsequently evaluating view synthesis on test views.

\myparagraph{pixelNeRF-DTU with finetuning and DS.} This variation is implemented similarly as \textit{pixelNeRF-DTU with finetuning}, but the pixelNeRF finetuning stage incorporates our depth supervision loss.

\subsection{Dataset Splits}
\myparagraph{NeRF Real-world.} We split each scene from NeRF Real into training views and test views. Because the number of views of each scene varies, we split every eighth image id into the test set (0, 8, 16, 24, \ldots) and construct training views that are evenly distributed over the remaining viewpoint id numbers. This setting gives us sufficient coverage to train different NeRF experiments. We create subsets from these training views of specific sizes to evaluate performance on few-input view synthesis.

\begin{table*}
\setlength{\tabcolsep}{7.5pt}
\center
\begin{tabular}{@{}l |  c c c | c c c | c c c }
\toprule
 & \multicolumn{3}{c}{PSNR$\uparrow$} & \multicolumn{3}{c}{SSIM$\uparrow$} & \multicolumn{3}{c}{LPIPS$\downarrow$}  \\
 \textbf{NeRF real-world} & 2-view & 5-view & 10-view & 2-view & 5-view & 10-view & 2-view & 5-view & 10-view\\
\midrule
NeRF & 13.52 & 18.37 & 22.29 & 0.39 & 0.49 & 0.53 & 0.55 & 0.51 & 0.53\\
metaNeRF-NeRF Real & 10.37 & 10.40 & 10.40 & 0.34 & 0.34 & 0.34 & 0.92 & 0.92 &  0.92 \\
pixelNeRF-NeRF Real & 11.56 & 11.59 & 11.61 & 0.36 & 0.37 & 0.37 & 0.87 & 0.87 & 0.86\\
\midrule
DS-NeRF  & \textbf{20.26} & \textbf{22.47} & \textbf{25.15} & \textbf{0.68} & \textbf{0.69} & \textbf{0.69} & \textbf{0.38} & \textbf{0.37} & \textbf{0.37}\\
\bottomrule
\end{tabular}
\caption{\textbf{View Synthesis on NeRF Real (4-fold cross-validation):} Because metaNeRF and pixelNeRF may perform better by training on sequences similar to the validation set, we perform 4-fold cross validation on 8 NeRF Real scenes. %
We find that training on NeRF Real scenes does not improve metaNerf/pixelNerf performance over training on DTU (see main paper), possibly because DTU-train has 88 scenes while a 4-fold NeRF Real training dataset has 6 scenes.
}

\lbltbl{nerf_real_supp}
\vspace{-0.2em}
\end{table*}
\myparagraph{Redwood 3d-scan.}
\label{sec:redwood}
We selected five test scene from the Redwood 3dscan dataset: \texttt{table}, \texttt{plant}, \texttt{chair}, \texttt{car}, and  \texttt{stool}.

Each scene is constructed using 15 frames of RGB and depth. These 15 frames are further sub-divided into training views and test views. We construct training sets with the following viewpoints [5, 11, 2, 8, 14, 1, 4, 7, 10, 13], truncating when working with a smaller subset (e.g. 2-views uses only 5 and 11). We evaluate the quality of view synthesis on the test views [0, 3, 6, 9, 12].

\subsection{Depth Error Evaluation}
\lblsec{depth_error_eva}
To evaluate the depth error of these different baselines, we need to first compute a reference depth of an input scene from test camera poses. 

\myparagraph{Depth evaluation on NeRF Real.} For a given set of training views from a scene, we use COLMAP's SfM algorithm to get sparse keypoints and camera poses. 
In addition, we run dense MVS on all training and test views to get a reference depth map from every view and test poses. To align dense MVS depth with the SfM keypoint depth obtained from the training views, we compute a scale $a$ and shift $b$ scalar that aligns the keypoint depth visible in a training view to its MVS depth. More specifically, we solve the following least squares optimization on those detected keypoints:

\begin{equation}
\min_{a, b} \sum_{p \in \mathcal{P}} ( a D_\mathrm{SfM}(p) - b - D_\mathrm{MVS}(p) )^2,
\label{eq:redwood_scale}
\end{equation}
where $\mathcal{P}$ is the set of detected and visible keypoints $p$ for a scene, $D_\mathrm{SfM}$ is the sparse depth we use for training, and $D_\mathrm{MVS}$ is the dense depth maps we use for evaluation.

Depth error can be computed by transforming the rendered depth $\hat{D}$ from a test camera $c$.

\begin{equation}
\mathrm{Err(D)} =  \left\lVert a \hat{D} - b - D_\mathrm{MVS} \right\rVert^2
\label{eq:redwood_error}
\end{equation}

Note that dense MVS depth is only used for depth evaluation. It is not used during training and test by any method.

\section{Additional Experiments}

\subsection{NeRF Real with pixelNeRF and metaNeRF}
While pixelNeRF and metaNeRF can achieve reasonable results when given sufficient pre-training data and a small train-test domain gap, many real-world applications cannot rely on the assumptions of similar test domain and sufficient training samples. We further highlight this by showing what happens to these baselines in such a scenario. To construct a more realistic setting for NeRF-based applications, we perform a 4-fold evaluation by splitting NeRF Real into $6$ training scenes and $2$ test scenes. The 4 test-splits are: fern-horn, flower-fortress, leaves-orchids, and room-trex.

For every split, we use the remaining $6$ training scenes to learn NeRF Real priors for pixelNeRF and metaNeRF and evaluate the view synthesis results after fine-tuning the model on the test scenes. We show these baseline results in \reftbl{nerf_real_supp} and find that meta-learning NeRF baselines struggle to properly leverage the priors observed during training.

For metaNeRF, we use the same ShapeNet setup described in \refsec{baseline}. We train on NeRF Real training scenes for $40\mathrm{K}$ steps and then finetune for $1\mathrm{K}$ steps on test scenes (\textit{metaNeRF-NeRF Real}). We also tried a slightly longer finetuning duration of $5\mathrm{K}$ steps, but found no significant improvement in results. We observed the following failure mode; within NeRF Real training, certain scenes were dominating the learned prior such that even with many iterations of fine-tuning on test-scenes, NeRF would only render cloudy looking textures from training scenes like \texttt{flowers} and \texttt{room}.

For pixelNeRF we train the network from scratch on the $6$ training scenes. Our training procedure uses 3 input views for each training iteration to finetune the encoder and decoder over $40\mathrm{K}$ iterations. Here we again find that the category-specific models struggle due to the lack of sufficient training data to learn these priors.

\subsection{MPI-based Experiments}

\begin{figure}
    \centering
    \includegraphics[width=.5\textwidth]{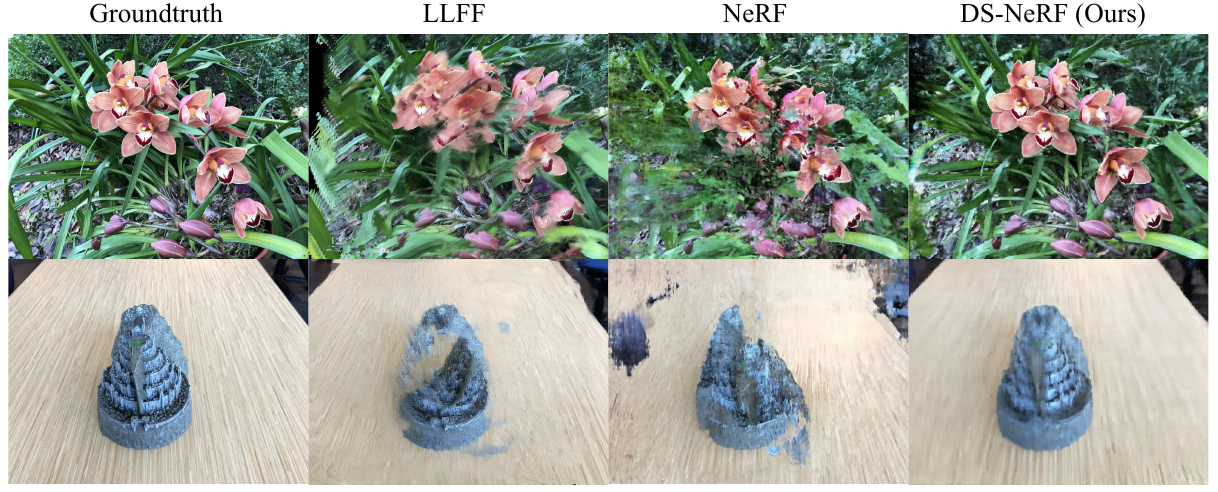}
    \caption{We visualize novel views on NeRF Real rendered by LLFF (MPI-based), NeRF, and DS-NeRF inferred from 2 input views. We find that despite LLFF being an MPI-based method, it still struggles to visually match the quality produced by DS-NeRF, agreeing with our quantitive experiments.}
    \lblfig{fig:llff}
\end{figure}

\begin{figure}
    \centering
    \includegraphics[width=.5\textwidth]{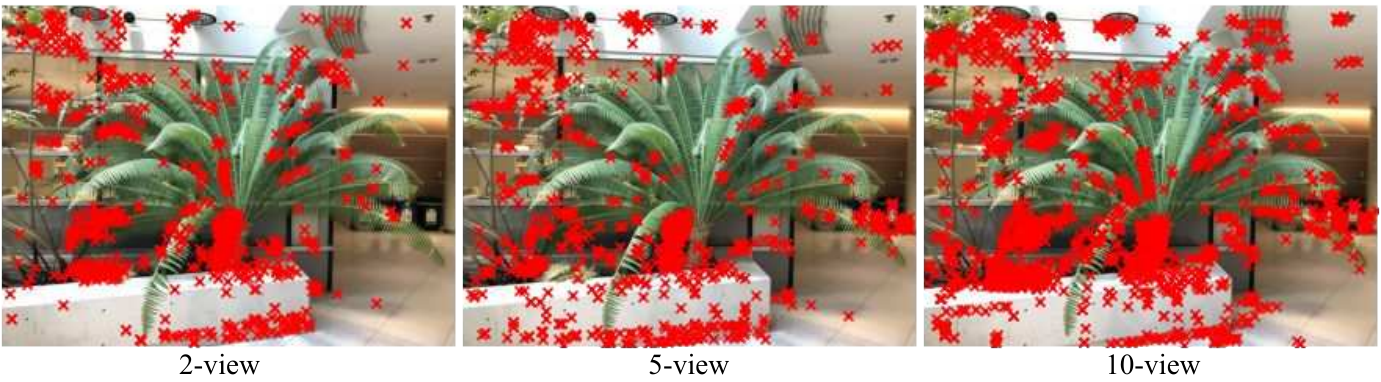}
    \caption{We visualize the keypoints detected by COLMAP with 2, 5, 10 input views to show the available depth supervision.}
    \lblfig{fig:kp}
\end{figure}

We consider Local Light Field Fusion (LLFF) \cite{mildenhall2019llff}, an MPI-based method, as a baseline on their dataset (referred to as NeRF Real). We have included qualitative comparisons on NeRF Real with LLFF in \reffig{fig:llff}. For quantitative comparison refer to Tab. 1 in the main paper. 

We find that while multi-plane images are a powerful representation, they have significant restrictions such as assuming
the renderable scene lies within a central frustrum and that depths from different views are related by a homography. As such applying LLFF to scenes like DTU and Redwood is impractical given the complex camera viewpoints compared to NeRF Real.

\subsection{Sparseness of keypoints}

We report the performance with varying sparseness of keypoint supervision. We ablate DS-NeRF by uniformly removing
detected keypoints while fixing the input to 5 views. When given (20\%, 50\%, and 100\%) of possible keypoints, DSNeRF’s
PSNR on test view is (21.5, 22.2, 22.6) respectively. Dropping out keypoints reasonably weakens the performance while still outperforming NeRF baseline (18.2).

We additionally consider how the quantity of keypoints degrade as fewer views are provided to COLMAP specifically. On average, there are 1615, 2172, 2621 keypoints for depth supervision per training view when given 2, 5, 10 views respectively. We also show qualitative examples in \reffig{fig:kp} for different number of input views.

\end{document}